\documentclass{article} 
\usepackage{iclr2027_conference,times}

\usepackage{amsmath,amsfonts,bm}

\def\eqref#1{equation~\ref{#1}}

\def\1{\bm{1}}

\DeclareMathAlphabet{\mathsfit}{\encodingdefault}{\sfdefault}{m}{sl}
\SetMathAlphabet{\mathsfit}{bold}{\encodingdefault}{\sfdefault}{bx}{n}

\usepackage{amsmath}
\usepackage{amssymb}
\usepackage{mathtools}
\usepackage{hyperref}
\usepackage{url}
\usepackage{graphicx}
\usepackage{booktabs}    
\usepackage{colortbl}    
\usepackage{xcolor}      
\usepackage{fontawesome5}
\usepackage{multirow}    

\usepackage{colortbl}

\newcommand{\ie}{\textit{i.e.}}

\usepackage{bm}

\usepackage[normalem]{ulem} 
\DeclareMathSizes{10}{8}{6}{5}
\usepackage{algorithm}
\usepackage{algorithmic}
\usepackage{CJKutf8}

\newcommand{\pstd}[1]{%
  \nobreak\hspace{0.06em}%
  \raisebox{-0.50ex}{%
    \textcolor{gray!70}{\scalebox{0.72}{$\pm #1$}}%
  }%
}

\title{\textbf{Adaptive Reward Routing:} Dynamic Multi-Reward Optimization for Joint Audio-Video Diffusion via Forward-Process RL}

\author{%
  Songlin Yang$^{1}$, Xiaotong Zhao$^{2}$, Jiacheng Zhang$^{3}$, Zhe Wang$^{1}$, Toyota Li$^{2}$, Eric Liu$^{2}$,\\ \textbf{Alan Zhao}$^{2}$, \textbf{Anyi Rao}$^{1}$ \\
  $^{1}$MMLab@HKUST, The Hong Kong University of Science and Technology \\
  $^{2}$Tencent Video, $^{3}$The University of Hong Kong \\
}

\iclrfinalcopy 
\begin{document}

\maketitle

\begin{abstract}
Multi-reward guided reinforcement learning (\ie, RL) offers a promising way to improve joint audio-video diffusion models along several objectives, including modality-specific quality, cross-modal semantic alignment, and temporal synchronization. Its effectiveness, however, depends on two quantities that change during training: \emph{where reward-driven updates should act, and how competing rewards should be coordinated}. Existing methods tend to rely on fixed routing and reward weights, failing to track evolving model functions. To address these limitations, we propose \textbf{Adaptive Reward Routing} to jointly adapt update locations and reward coordination during forward-process RL (\ie, DiffusionNFT) of joint audio-video diffusion models. Our method consists of two components. (i) Cross-Modal Influence-Guided Routing (Localizing Updates): We use bidirectional cross-attention responses as an efficient proxy for evolving cross-modal influence, dynamically reweighting token-aware losses and scaling gradients across cross-modal layers without additional model interventions. (ii) Preference-Preserving Modality-Aware Reweighting (Coordinating Rewards): We preserve predefined weights as preference priors and use branch-specific reward-gradient interactions as residual corrections after warm-up. This resolves evolving conflicts without letting dominant rewards suppress weak but essential objectives. Extensive experiments demonstrate consistent improvements in modality quality, semantic consistency, and audio-video synchronization over strong RL baselines. Ablations and mechanism analyses further validate the complementary benefits of adaptive update routing and reward coordination.
\end{abstract}

\section{Introduction}

Recent advances in joint audio-video diffusion models~\citep{ltx2} have enabled the generation of visual and audio content from text prompts. However, high-quality joint generation must simultaneously satisfy modality-specific visual and audio quality, cross-modal semantic alignment, and temporal synchronization, which are difficult to capture with a single supervised objective. Reward-guided diffusion reinforcement learning (RL), including GRPO-based methods~\citep{grpo,flow-grpo} and DiffusionNFT~\citep{diffusionnft}, therefore provides a promising paradigm by expressing these requirements through multiple reward signals.

\begin{figure}[t]
    \centering
    \includegraphics[width=\linewidth]{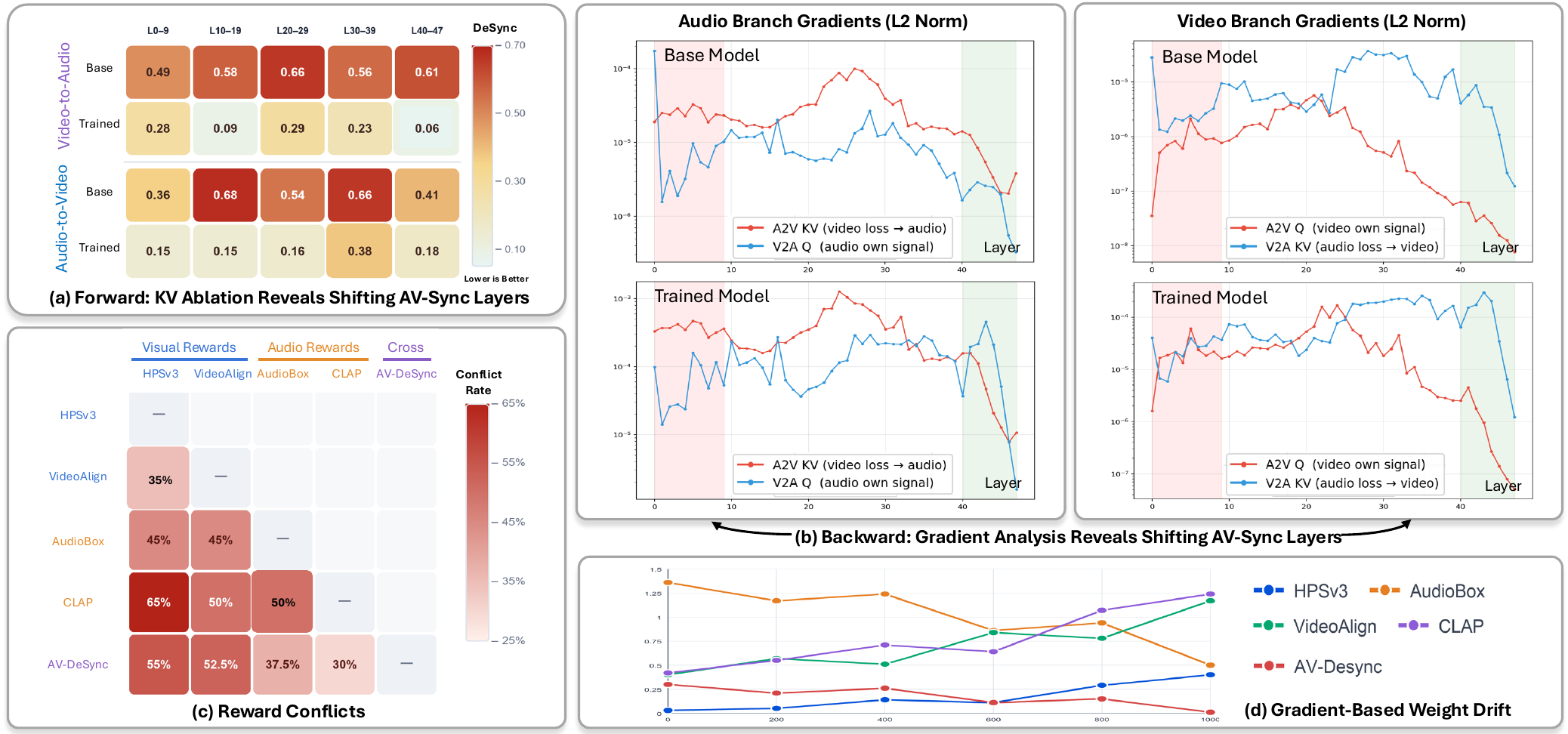}
    \vspace{-0.4cm}
    \caption{\textbf{Joint audio-video post-training requires adaptive update routing and reward coordination.} \textbf{(a) Forward KV Ablation.} We ablate cross-modal K/V outputs by layer range and measure DeSync (lower is better; larger degradation indicates greater synchronization importance). Divergent base/trained profiles reveal shifting synchronization-critical layers. \textbf{(b) Backward Gradient Analysis.} We measure layer-wise $\ell_2$ norms of Q (within-branch) and K/V (cross-modal) gradients in both branches; shifted base/trained peaks show evolving cross-modal gradient paths. \textbf{(c) Reward Conflicts.} Rewards conflict when their per-prompt normalized preferences have opposite signs, with larger off-diagonal values indicating more disagreement. The conflicts show that fixed branch assignment is insufficient. \textbf{(d) Gradient-Based Weight Drift.} We track weights derived solely from reward-gradient geometry, where near-zero trajectories indicate objective suppression. The vanishing AV-DeSync weight shows that geometry-only weighting can discard essential objectives. Together, these results motivate dynamic routing and preference-anchored reward coordination.}
    \label{fig:teaser}
\end{figure}

However, reward-guided RL of joint audio-video diffusion models remains challenging, as it involves dynamic multi-reward optimization along two coupled dimensions: \emph{where should reward-driven updates act, and how should multiple rewards be coordinated?} \textbf{(i) Dynamic Reward Routing: Where to Optimize.} Joint audio-video models contain modality-specific branches coupled through cross-attention. Reward routers first determine the responsible modality branches, while the resulting branch-level updates must be localized across tokens and cross-modal interaction layers. OmniNFT~\citep{omninft} recognizes these but fixes its layer routing based on the base model. However, our probing of the base and OmniNFT-trained checkpoints in Fig.~\ref{fig:teaser}(a) and (b) shows that cross-modal functions and gradient flows evolve during fine-tuning, making static routing progressively stale. \textbf{(ii) Dynamic Reward Coordination: How to Balance.} Rewards frequently disagree on the same sample, as shown in Fig.~\ref{fig:teaser}(c), and their appropriate balance changes throughout optimization. GDPO~\citep{gdpo} normalizes each reward but combines them through fixed weights, leaving conflicts unadapted. MARBLE~\citep{marble} adjusts weights using gradient geometry, but its coefficients reflect gradient compatibility rather than importance aligned with user preferences, as shown in Fig.~\ref{fig:teaser}(d). Effective post-training therefore requires conflict-aware adaptation anchored by user-defined priorities.

Together, these challenges call for an approach that adapts both reward coordination and update routing as the model evolves. We therefore propose \textbf{Adaptive Reward Routing} for forward-process RL (\ie, DiffusionNFT~\citep{diffusionnft}) of joint audio-video diffusion models. It has two components. \textbf{(i) Cross-Modal Influence-Guided Routing} uses bidirectional cross-attention responses to locate reward-driven updates. Layer aggregation yields token weights that emphasize cross-modally influential locations, while token aggregation yields layer scales that preserve gradients through influential cross-modal pathways. \textbf{(ii) Preference-Preserving Modality-Aware Reweighting} estimates reward conflicts within the branch responsible for each objective. After warm-up, it uses the resulting coefficients as residual corrections to predefined reward weights, adapting to changing conflicts without overriding user priorities.

Our experiments establish three findings. First, Adaptive Reward Routing consistently improves modality quality, semantic alignment, and audio-video synchronization of joint audio-video diffusion models. Second, controlled ablations verify the complementary contributions of token- and layer-level routing, branch-aware conflict estimation, residual preference correction, and warm-up. Third, mechanism analyses with direct path interventions confirm that the cross-attention response proxy identifies functionally important layers and tokens, while routes frozen at initialization become stale as training progresses.

\textbf{Contributions.} (i) We formulate joint audio-video diffusion RL as dynamic multi-reward optimization over two coupled dimensions (\ie, where modality-conditioned reward updates should act and how multiple rewards should be coordinated) and empirically reveal the limitations of static solutions. (ii) We propose Adaptive Reward Routing, which unifies cross-modal response-guided token/layer localization with preference-preserving, modality-aware reward coordination. (iii) We provide comprehensive comparisons, ablations, and mechanism analyses demonstrating that adapting both dimensions enables more stable and effective joint audio-video post-training.

\section{Related Work}

\paragraph{Joint Audio-Video Generation.}
Video generation~\cite{shotverse,context} has progressed from image diffusion models with temporal modules~\citep{SVD,animatediff} to large diffusion Transformers~\citep{hunyuanvideo}, with flow matching~\citep{wan,ltx-video} and compressed latents improving efficiency~\citep{evalverse}. Joint audio-video systems connect pretrained experts through cross-modal projections~\citep{universe-1}, use unified diffusion Transformers~\citep{javisdit,javisdit++}, or couple separate streams through bidirectional cross-attention~\citep{ltx2}. These heterogeneous branches enable mutual conditioning but make reward responsibility and gradient routing less obvious than in a single-stream model.

\paragraph{Reinforcement Learning for Diffusion Models.}
GRPO~\citep{grpo} estimates relative advantages without a critic. Flow-GRPO~\citep{flow-grpo} and DanceGRPO~\citep{dancegrpo} extend online optimization to flow-based generation through stochastic sampling. DiffusionNFT~\citep{diffusionnft} instead optimizes the forward process using implicit positive and negative policies. OmniNFT~\citep{omninft} adds modality-wise credit assignment for joint audio-video generation. We retain its forward-process formulation but replace fixed routing rules with token- and layer-level routes recomputed from the current model.

\paragraph{Multi-Reward Optimization.}
Fixed scalarization cannot react to changing conflicts. GDPO~\citep{gdpo} preserves reward-specific signals through decoupled normalization, but still uses predefined aggregation weights. Multi-task methods instead seek common descent directions~\citep{mgda,multi-task}, project conflicting gradients~\citep{gradient}, or optimize local agreement~\citep{conflict}. MARBLE~\citep{marble} adapts this idea to diffusion RL. Because gradient compatibility alone does not encode objective importance or modality responsibility, we estimate conflicts within each modality branch and use them as residuals to preference priors.

\section{Problem Formulation and Preliminaries}

We study reward-guided post-training of joint audio-video diffusion models (\ie, LTX-2~\citep{ltx2}) under DiffusionNFT~\citep{diffusionnft}, which can be formulated as multi-modal, multi-reward forward-process reinforcement learning. We use $m\in\mathcal M=\{v,a\}$ for modality, $n$ for rollout sample, $k$ for reward, $i$ for token, $l$ for Transformer block, and $t$ for flow-matching timestep.

\paragraph{Joint Audio-Video Flow Matching.}

LTX-2~\citep{ltx2} uses separate audio and video streams under a shared timestep. Each latent follows the standard linear interpolation $x_t^m=(1-t)x_0^m+t x_1^m$, with $x_1^m\sim\mathcal N(0,I)$, and the model predicts the two velocity fields jointly. The streams exchange information through bidirectional cross-attention:
\begin{equation}
    o_{a\rightarrow v}^{l,t}
=\operatorname{Attn}\!\left(Q_v(h_v^{l,t}),K_a(h_a^{l,t}),V_a(h_a^{l,t})\right),\quad
    o_{v\rightarrow a}^{l,t}
=\operatorname{Attn}\!\left(Q_a(h_a^{l,t}),K_v(h_v^{l,t}),V_v(h_v^{l,t})\right).
\label{eq:bidirectional_cross_attention}
\end{equation}
The gated audio-to-video (A2V) and video-to-audio (V2A) outputs are added to the video and audio streams, respectively.

\paragraph{Diffusion Forward-Process Reinforcement Learning.}

DiffusionNFT~\citep{diffusionnft} constructs implicit positive and negative policies from the updated and trainable velocity predictors:
\begin{equation}
    v_\theta^+=(1-\beta)v^{\mathrm{updated}}+\beta v_\theta,
    \qquad
    v_\theta^-=(1+\beta)v^{\mathrm{updated}}-\beta v_\theta.
\label{eq:nft_policies}
\end{equation}
For each prompt, the updated policy generates a group of $N$ samples. The reward of sample $n$ is converted to a group-relative advantage,
\begin{equation}
    A^{(n)}=\frac{R^{(n)}-\mu_R}{\sigma_R+\varepsilon},
    \qquad
    r^{(n)}=\frac{1}{2}+\frac{1}{2}\operatorname{clip}\!\left(\frac{A^{(n)}}{A_{\max}},-1,1\right),
\label{eq:nft_advantage_probability}
\end{equation}
where $\mu_R$ and $\sigma_R$ are computed within the rollout group. Thus $r^{(n)}>1/2$ favors the positive policy, whereas $r^{(n)}<1/2$ favors the negative policy. The resulting objective is
\begin{equation}
    \mathcal L_{\mathrm{NFT}}
    =\mathbb E_{n,t}\!\left[
    r^{(n)}\|v_\theta^+(x_t^{(n)},c,t)-u^{(n)}\|_2^2
    +(1-r^{(n)})\|v_\theta^-(x_t^{(n)},c,t)-u^{(n)}\|_2^2
    \right].
\label{eq:nft_loss}
\end{equation}
This advantage requires no learned value function: it states only whether a sample performs above or below its peers for the same prompt.

\paragraph{Multi-Reward Optimization.}

Let $\mathcal K=\mathcal K_v\cup\mathcal K_a\cup\mathcal K_c$ denote video, audio, and cross-modal rewards. Eq.~\ref{eq:nft_advantage_probability} is applied independently to each reward, producing $A_k^{(n)}$. GDPO~\citep{gdpo} combines them using predefined weights, $A_{\mathrm{GDPO}}^{(n)}=\sum_k\omega_k^{\mathrm{prior}}A_k^{(n)}$. MARBLE~\citep{marble} instead chooses simplex weights that minimize the norm of the weighted sum of normalized reward gradients. GDPO therefore preserves explicit preferences but cannot adapt to conflicts, while MARBLE adapts to local gradient geometry but does not encode preference or modality responsibility.

\section{Method: Adaptive Reward Routing}

\subsection{Overview}

\begin{figure}[t]
    \centering
    \includegraphics[width=\linewidth]{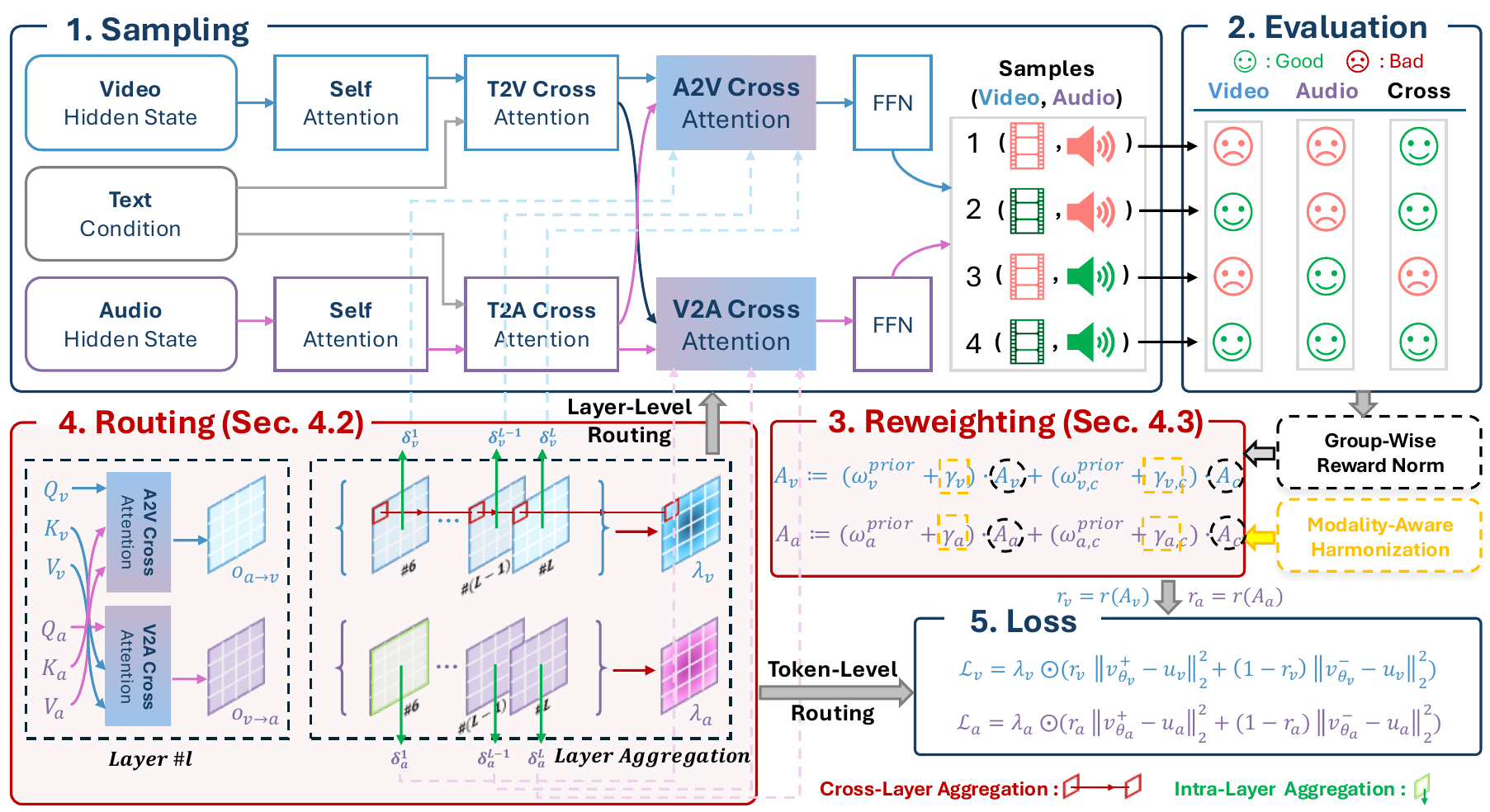}

    \caption{\textbf{Overview of Adaptive Reward Routing.} Our framework adapts reward-driven optimization at four levels: reward reweighting, modality-branch assignment, token-level credit allocation, and layer-wise cross-modal gradient routing.}
    \label{fig:framework}

\end{figure}

We propose \textbf{Adaptive Reward Routing}, a forward-process RL framework that adapts both reward priorities and routing locations, as shown in Fig.~\ref{fig:framework}. It contains two components. First, \textbf{Cross-Modal Influence-Guided Routing} (Sec.~\ref{method:routing}) determines where the update should act by adapting token weights and layer-wise cross-modal gradient flow. Second, \textbf{Preference-Preserving Modality-Aware Reweighting} (Sec.~\ref{method:weighting}) determines how rewards should be combined within the video and audio branches. As shown in Algorithm~\ref{alg:adaptive_reward_routing}, the complete optimization flow is
\begin{equation}
    \{A_k\}
    \xrightarrow[\text{(Sec.~4.3)}]
               {\text{reward reweighting}}
    \{\omega_{m,k}A_k\}
    \xrightarrow[\text{(Sec.~4.3)}]
               {\text{branch routing}}
    (A_v,A_a)
    \xrightarrow[\text{(Sec.~4.2)}]
               {\text{token routing}}
    \mathcal{L}
    \xrightarrow[\text{(Sec.~4.2)}]
               {\text{layer routing}}
    \nabla_\theta\mathcal{L}.
\label{eq:routing_overview}
\end{equation}
Intuitively, reward reweighting decides how strongly each objective contributes, branch routing assigns objectives to target modalities, token routing selects where each modality loss is emphasized, and layer routing controls how the resulting gradient crosses modality boundaries.

\begin{algorithm}[t]
\caption{Adaptive Reward Routing}
\label{alg:adaptive_reward_routing}
\footnotesize
\begin{algorithmic}[1]
  \REQUIRE Policy $v_\theta$, updated policy $v^{\mathrm{updated}}$, and smoothed coefficients $\bar\gamma_m$
  \FOR{each training round $e$}
      \STATE Generate a group of samples with $v^{\mathrm{updated}}$ and collect pre-gate A2V/V2A responses during sampling
      \STATE Evaluate $\{R_k^{(n)}\}$ and independently normalize each reward to obtain $\{A_k^{(n)}\}$
      \STATE Compute token weights $\{\lambda_{m,i}^{(n)}\}$ using Eq.~\ref{eq:token_routing_weight} and layer routes $\{\alpha_m^l\}$ using Eq.~\ref{eq:layer_routing_score}
      \STATE Compute $\{\omega_{m,k}\}$ from the cached $\bar\gamma_m$ using Eq.~\ref{eq:residual_reweighting}
      \STATE Construct $A_m^{(n)}$ using Eq.~\ref{eq:routed_advantages} and set $r_m^{(n)}=r(A_m^{(n)})$
      \IF{$e\ge e_{\mathrm{warm}}$ and $e$ is a refresh round}
          \STATE Probe reward gradients through their responsible branches with uniform token weights
          \STATE Solve the branch-wise MARBLE problem and update $\bar\gamma_m$ for the next round
      \ENDIF
      \STATE Compute $\ell_{m,i}^{(n)}$, $\mathcal L_m^{\mathrm{policy}}$, and $\mathcal L(\theta)$ using Eqs.~\ref{eq:modality_token_loss}--\ref{eq:full_objective}
      \STATE Backpropagate through the routed KV paths in Eq.~\ref{eq:kv_routing} and update $\theta$
      \STATE Update $v^{\mathrm{updated}}$ according to the DiffusionNFT policy-update schedule
  \ENDFOR
\end{algorithmic}
\end{algorithm}

\subsection{Cross-Modal Influence-Guided Routing}

\label{method:routing}

A direct measure of directional influence would disable A2V or V2A and compare the velocity predictions. Repeating this intervention during training would require extra model evaluations. We instead use a quantity already produced by the forward pass: the pre-gate response of the corresponding cross-attention path. For target token $i$,
\begin{equation}
    d_{v,i}^{l,t}=\|o_{a\rightarrow v,i}^{l,t}\|_2,
    \qquad
    d_{a,i}^{l,t}=\|o_{v\rightarrow a,i}^{l,t}\|_2.
\label{eq:cross_modal_proxy}
\end{equation}
These directional responses are collected over an intermediate-to-late denoising window and detached before policy optimization. Sec.~\ref{sec:proxy_validation} validates their relationship to direct interventions.

\paragraph{Token-Level Routing.}
For each target token, we average its responses over the selected timesteps and cross-modal blocks. After percentile-clipped min--max normalization ($\operatorname{Norm}_{99}$), the score becomes a positive loss weight:
\begin{equation}
    \lambda_{m,i}=1+(\lambda_{\max}-1)\operatorname{Norm}_{99}\!\left(
    \frac{1}{|\mathcal B||\mathcal T|}\sum_{l\in\mathcal B}\sum_{t\in\mathcal T}d_{m,i}^{l,t}
    \right).
\label{eq:token_routing_weight}
\end{equation}

Audio responses are normalized globally, while video responses are normalized within each frame to prevent frame-level magnitude differences from dominating the weights. These weights are applied to the token-level negative-aware loss in Sec.~\ref{method:objective}.

\paragraph{Layer-Level Routing.}
For each layer, we instead average the same response over tokens and selected timesteps. Let $\widetilde\delta_m^l$ denote this layer score after min--max normalization across blocks. We convert it to a soft detachment coefficient
\begin{equation}
    \alpha_m^l=(1-\widetilde\delta_m^l)^{1/\tau}.
\label{eq:layer_routing_score}
\end{equation}
For a source key or value tensor $X\in\{K,V\}$, the routed representation is
\begin{equation}
    \widetilde X_{\bar m\rightarrow m}^{l,t}
    =\alpha_m^l\operatorname{sg}(X_{\bar m}^{l,t})+(1-\alpha_m^l)X_{\bar m}^{l,t}.
\label{eq:kv_routing}
\end{equation}
This operation leaves the forward value unchanged but scales its backward gradient by $1-\alpha_m^l$. Strongly influential layers retain more gradient, while weakly coupled layers are increasingly detached. A2V and V2A are routed independently.

\subsection{Preference-Preserving Modality-Aware Reweighting}
\label{method:weighting}

\paragraph{Motivation.} Predefined reward weights express what the user wants, but they cannot react to reward conflicts. Gradient-based coefficients react to conflicts, but may suppress a weak objective because its early gradient is noisy or incompatible. We combine the two rather than choosing one.

\paragraph{Implementation.} Each reward is probed only through the branch it supervises: video and audio rewards use their respective branches, while cross-modal rewards use both. MARBLE then produces a conflict-aware coefficient $\gamma_{m,k}$ within each branch. Token routing is disabled during these probes so that the measured geometry is not biased by the current token weights. After warm-up, the smoothed coefficient provides a residual correction to the prior:
\begin{equation}
    \omega_{m,k}=
    \begin{cases}
        \omega_{m,k}^{\mathrm{prior}}, & e<e_{\mathrm{warm}},\\
        (1-\kappa)\omega_{m,k}^{\mathrm{prior}}+\kappa C_m\bar\gamma_{m,k}, & e\ge e_{\mathrm{warm}}.
    \end{cases}
\label{eq:residual_reweighting}
\end{equation}
Here $C_m$ rescales the simplex coefficients to preserve the total prior weight within branch $m$, and $\bar\gamma_m\leftarrow\rho\bar\gamma_m+(1-\rho)\gamma_m^*$ smooths successive estimates. The prior therefore sets a nonzero floor, while the residual term adapts to current conflicts.

\subsection{Training Objective}
\label{method:objective}

The adaptive reward weights first produce a separate advantage for each modality branch:
\begin{equation}
    A_m^{(n)}
    =
    \sum_{k\in\mathcal K_m\cup\mathcal K_c}
    \omega_{m,k}A_k^{(n)},
    \qquad m\in\{v,a\}.
\label{eq:routed_advantages}
\end{equation}
Cross-modal rewards are included in both branches. We then map each branch advantage to an optimality probability $r_m^{(n)}=r(A_m^{(n)})$ using Eq.~\ref{eq:nft_advantage_probability}. For token $i$ of sample $n$, the negative-aware loss is
\begin{equation}
    \ell_{m,i}^{(n)}
    =
    r_m^{(n)}
    \frac{\|v_{\theta,m,i}^{+}-u_{m,i}\|_2^2}{w_m^{+,(n)}+\varepsilon}
    +
    \bigl(1-r_m^{(n)}\bigr)
    \frac{\|v_{\theta,m,i}^{-}-u_{m,i}\|_2^2}{w_m^{-,(n)}+\varepsilon}.
\label{eq:modality_token_loss}
\end{equation}
Here $w_m^{\pm,(n)}$ is the detached mean absolute residual of the corresponding policy, averaged over all tokens and feature dimensions of modality $m$. The token routing weights then form the modality loss
\begin{equation}
    \mathcal L_m^{\mathrm{policy}}
    =
    \mathbb E_n\!\left[
    \frac{
        \sum_{i\in\mathcal I_m}
        \lambda_{m,i}^{(n)}\ell_{m,i}^{(n)}
    }{
        \sum_{i\in\mathcal I_m}\lambda_{m,i}^{(n)}
    }
    \right].
\label{eq:weighted_policy_loss}
\end{equation}
Finally, we combine the two branches and regularize them toward the fixed reference policy:
\begin{equation}
    \mathcal L(\theta)
    =
        \sum_{m\in\mathcal M}\mathcal L_m^{\mathrm{policy}}
    +
    \lambda_{\mathrm{KL}}
        \sum_{m\in\mathcal M}\mathcal L_{\mathrm{KL},m}(\theta).
\label{eq:full_objective}
\end{equation}

\section{Experiments}
\label{sec:experiments}

\subsection{Experimental Setup}
\label{sec:experimental_setup}

\paragraph{Backbones and Training Data.} We evaluate Adaptive Reward Routing on two joint audio-video diffusion backbones, LTX-2 (19B) and LTX-2.3 (22B)~\citep{ltx2}. Both models employ separate audio and video streams connected through bidirectional cross-attention, making them suitable for studying adaptive reward localization across modalities, tokens, and layers. For reward-guided post-training, we use 19,487 audio-video prompts collected from a VGGSound-derived~\citep{vggsound} corpus. Each record contains modality-specific audio and video descriptions together with a joint audio-video prompt.

\paragraph{Reward Models.} Following the multi-objective evaluation dimensions of joint audio-video generation, we optimize five complementary reward signals: (i) Video Quality: VideoAlign~\citep{videoalign} and HPSv3~\citep{hpsv3}; (ii) Audio Quality: AudioBox Aesthetics~\citep{audiobox_aesthetics}; (iii) Text-Audio Alignment: CLAP~\citep{CLAP}; and (iv) Audio-Video Synchronization: DeSync~\citep{synchformer}, which we convert into a higher-is-better AV-DeSync during training, while reporting the original lower-is-better DeSync metric at evaluation time.

\paragraph{Baselines.}
We compare against complementary reference settings that cover the principal dimensions of multimodal, multi-reward optimization. (i) No Post-Training: the pretrained backbone establishes the performance before reward-guided optimization. (ii) Fixed Reward Coordination: GDPO~\citep{gdpo} independently normalizes reward-wise advantages but aggregates them using fixed weights. (iii) Conflict-Aware Reward Coordination: MARBLE~\citep{marble} dynamically adjusts reward weights according to global gradient conflicts, without modality-specific routing. (iv) Static Multimodal Routing: OmniNFT~\citep{omninft}, designed specifically for DiffusionNFT-based joint audio-video post-training, introduces modality-wise credit assignment, layer-wise gradient surgery, and region-wise reweighting, but fixes its routing strategy using the base model. OmniNFT* denotes the checkpoint released by the original authors. 

\paragraph{Evaluation.} We evaluate on the full JavisBench benchmark~\citep{javisdit++}, which contains 10,140 prompts spanning diverse audio-video generation scenarios. Generated outputs are normalized to the benchmark protocol of four seconds, 24 FPS, and 16-kHz audio. We report four groups of metrics: (i) AV-Quality~\citep{videoalign}, including Visual Quality (VQ) and Audio Quality (AQ); (ii) Text-Consistency, including Text-Video and Text-Audio ImageBind similarity (TV-IB and TA-IB)~\citep{imagebind}, CLIP~\citep{clip}, and CLAP; (iii) AV-Consistency~\citep{imagebind}, including AV-IB and AVHScore; and (iv) AV-Synchrony, including JavisScore~\citep{javisdit} and DeSync.

\begin{figure}
    \centering
    \includegraphics[width=\linewidth]{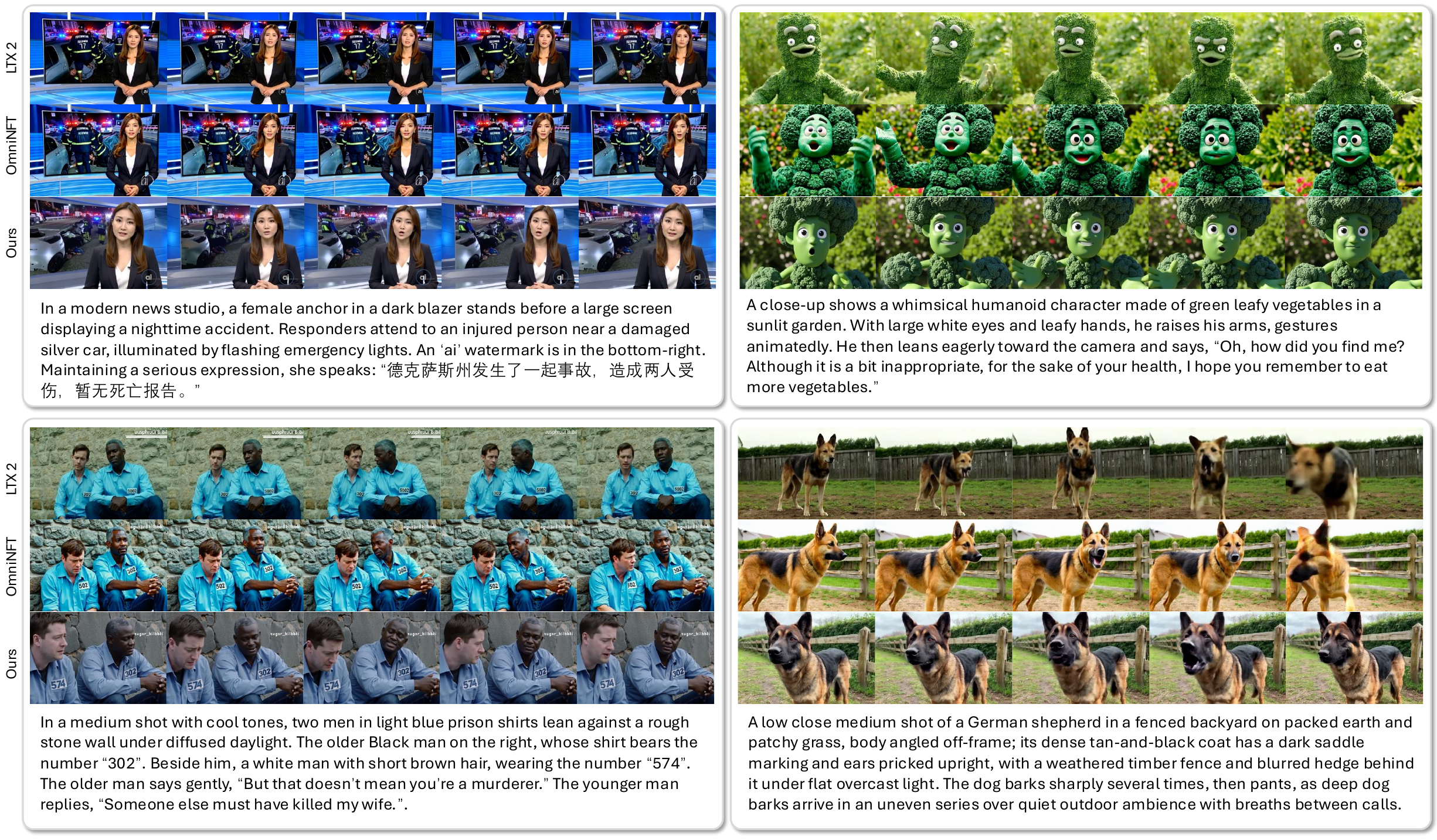}

    \caption{\textbf{Qualitative comparison of joint audio-video generation.} For each prompt, we show five temporally ordered frames generated by LTX-2, OmniNFT, and Ours. The examples cover Chinese news delivery, stylized English speech, two-speaker dialogue, and dog barking. Our method better preserves the requested subjects and scene composition while reducing identity and appearance drift throughout the generated sequence.}
    \label{fig:video}

\end{figure}

\begin{table}[t]
  \centering
  \caption{\textbf{Main results on JavisBench.} (Mean ± std over 3 seeds.)}
  \label{tab:javisbench_results}
  \resizebox{\textwidth}{!}{
  \begin{tabular}{ll ll llll ll ll}
  \toprule
  \multirow{2}{*}{Backbone}
  & \multirow{2}{*}{Method}
  & \multicolumn{2}{c}{AV-Quality}
  & \multicolumn{4}{c}{Text-Consistency}
  & \multicolumn{2}{c}{AV-Consistency}
  & \multicolumn{2}{c}{AV-Synchrony} \\
  \cmidrule(lr){3-4}
  \cmidrule(lr){5-8}
  \cmidrule(lr){9-10}
  \cmidrule(lr){11-12}
  &
  & VQ$\uparrow$
  & AQ$\uparrow$
  & TV-IB$\uparrow$
  & TA-IB$\uparrow$
  & CLIP$\uparrow$
  & CLAP$\uparrow$
  & AV-IB$\uparrow$
  & AVHScore$\uparrow$
  & JavisScore$\uparrow$
  & DeSync$\downarrow$ \\
  \midrule

  \multirow{6}{*}{LTX-2}
  & Base Model
  & 1.883
  & 5.201
  & 0.265
  & 0.143
  & 0.312
  & 0.358
  & 0.180
  & 0.177
  & 0.153
  & 0.604 \\

  & + GDPO
  & 2.722\pstd{0.0136}
  & 5.450\pstd{0.0217}
  & 0.261\pstd{0.0020}
  & 0.138\pstd{0.0018}
  & 0.312\pstd{0.0014}
  & 0.347\pstd{0.0031}
  & 0.174\pstd{0.0025}
  & 0.175\pstd{0.0022}
  & 0.155\pstd{0.0020}
  & 0.671\pstd{0.0104} \\

  & + MARBLE
  & 2.384\pstd{0.0124}
  & 5.100\pstd{0.0193}
  & 0.265\pstd{0.0016}
  & 0.138\pstd{0.0019}
  & 0.311\pstd{0.0013}
  & 0.365\pstd{0.0028}
  & 0.182\pstd{0.0023}
  & 0.182\pstd{0.0026}
  & 0.158\pstd{0.0021}
  & 0.618\pstd{0.0092} \\

  & + OmniNFT
  & 3.136\pstd{0.0108}
  & \underline{5.614}\pstd{0.0176}
  & \underline{0.265}\pstd{0.0018}
  & 0.145\pstd{0.0016}
  & 0.312\pstd{0.0011}
  & 0.416\pstd{0.0024}
  & \underline{0.222}\pstd{0.0019}
  & \underline{0.219}\pstd{0.0021}
  & \underline{0.195}\pstd{0.0017}
  & 0.390\pstd{0.0075} \\

  & + OmniNFT*
  & \underline{3.278}
  & 5.609
  & 0.254
  & \underline{0.165}
  & \textbf{0.314}
  & \underline{0.422}
  & 0.220
  & 0.219
  & 0.193
  & \underline{0.378} \\

  & + Ours
  & \textbf{3.336}\pstd{0.0094}
  & \textbf{5.868}\pstd{0.0158}
  & \textbf{0.268}\pstd{0.0014}
  & \textbf{0.167}\pstd{0.0013}
  & \underline{0.314}\pstd{0.0011}
  & \textbf{0.425}\pstd{0.0020}
  & \textbf{0.235}\pstd{0.0017}
  & \textbf{0.234}\pstd{0.0019}
  & \textbf{0.206}\pstd{0.0015}
  & \textbf{0.341}\pstd{0.0062} \\

  \midrule

  \multirow{6}{*}{LTX-2.3}
  & Base Model
  & 2.032
  & 5.218
  & 0.271
  & 0.151
  & 0.308
  & 0.387
  & 0.205
  & 0.202
  & 0.175
  & 0.504\\

  & + GDPO
  & 2.929\pstd{0.0142}
  & 5.251\pstd{0.0225}
  & \underline{0.272}\pstd{0.0021}
  & 0.144\pstd{0.0020}
  & 0.309\pstd{0.0015}
  & 0.376\pstd{0.0033}
  & 0.218\pstd{0.0027}
  & 0.199\pstd{0.0025}
  & 0.178\pstd{0.0022}
  & 0.560\pstd{0.0113} \\

  & + MARBLE
  & 2.582\pstd{0.0130}
  & 5.176\pstd{0.0206}
  & 0.271\pstd{0.0018}
  & 0.147\pstd{0.0017}
  & 0.309\pstd{0.0014}
  & 0.394\pstd{0.0029}
  & 0.217\pstd{0.0022}
  & 0.207\pstd{0.0028}
  & 0.182\pstd{0.0020}
  & 0.496\pstd{0.0098} \\

  & + OmniNFT
  & 3.489\pstd{0.0115}
  & 5.693\pstd{0.0181}
  & 0.271\pstd{0.0016}
  & 0.163\pstd{0.0015}
  & \textbf{0.316}\pstd{0.0010}
  & 0.449\pstd{0.0025}
  & 0.250\pstd{0.0021}
  & 0.238\pstd{0.0023}
  & \underline{0.224}\pstd{0.0018}
  & 0.369\pstd{0.0079} \\

  & + OmniNFT*
  & \underline{3.537}
  & \underline{5.702}
  & 0.260
  & \underline{0.174}
  & 0.311
  & \underline{0.456}
  & \underline{0.252}
  & \underline{0.249}
  & 0.221
  & \underline{0.335} \\

  & + Ours
  & \textbf{3.599}\pstd{0.0089}
  & \textbf{5.979}\pstd{0.0151}
  & \textbf{0.274}\pstd{0.0013}
  & \textbf{0.177}\pstd{0.0012}
  & \underline{0.314}\pstd{0.0011}
  & \textbf{0.460}\pstd{0.0019}
  & \textbf{0.267}\pstd{0.0016}
  & \textbf{0.266}\pstd{0.0017}
  & \textbf{0.236}\pstd{0.0014}
  & \textbf{0.302}\pstd{0.0058} \\

  \bottomrule
  \end{tabular}
  }

\end{table}

\subsection{Main Results and Training Dynamics}
\label{sec:main_results}

\begin{figure}
    \centering
    \includegraphics[width=\linewidth]{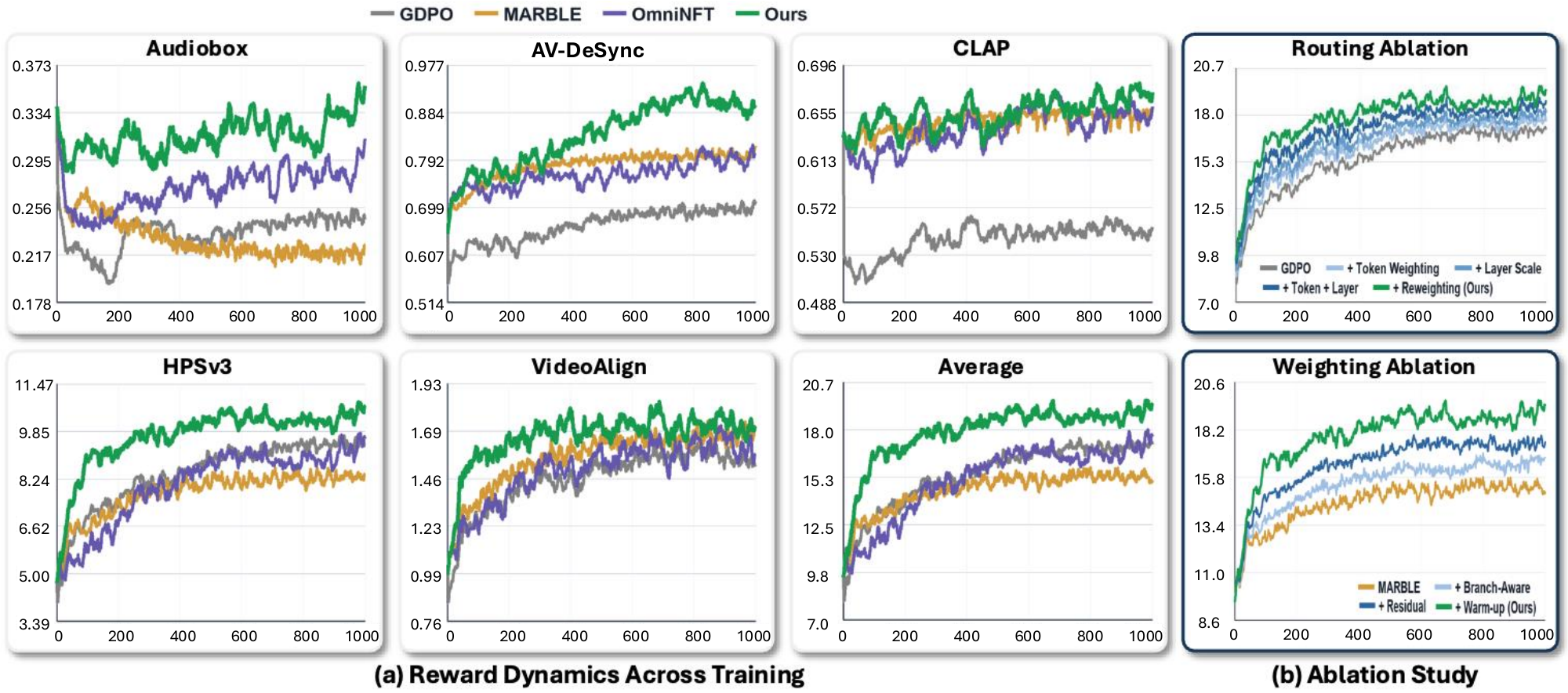}

    \caption{\textbf{Reward dynamics and component ablations.} (a) Training trajectories of five individual rewards (AudioBox, AV-DeSync, CLAP, HPSv3, and VideoAlign) and their average for GDPO, MARBLE, OmniNFT, and Ours. Average denotes the arithmetic mean of the five normalized rewards. (b) Ablation studies of the routing and weighting designs. Starting from GDPO or MARBLE, respectively, each added component yields progressive gains, while the complete method achieves the strongest overall performance.}
    \label{fig:reward}

\end{figure}

Fig.~\ref{fig:video}, Tab.~\ref{tab:javisbench_results}, and Fig.~\ref{fig:reward} summarize the generation quality, benchmark performance, and optimization behavior of our method, respectively. \textbf{(i) Qualitative Results.} Fig.~\ref{fig:video} covers diverse audio-video scenarios, including multilingual speech, a stylized speaking character, a two-speaker exchange, and animal vocalization. LTX-2 exhibits noticeable subject and appearance drift, particularly in the character and animal examples, while OmniNFT improves prompt fidelity but retains temporal inconsistencies. Our method maintains more stable identities and scene structures while preserving the visual actions associated with speech, dialogue, and barking. \textbf{(ii) Quantitative Results.} As shown in Tab.~\ref{tab:javisbench_results}, our method achieves the strongest overall performance on both LTX-2 and LTX-2.3, obtaining the best result on nine of the ten metrics under each backbone. For each backbone, GDPO, MARBLE, OmniNFT, and Ours are independently trained with three random seeds under the same data, LoRA, and optimization budgets, and the table reports their arithmetic means. The Base Model and OmniNFT* are fixed checkpoints evaluated under the same generation and evaluation protocol, with OmniNFT* denoting the checkpoint released by its original authors. GDPO improves visual quality but degrades several audio and synchronization metrics, revealing the imbalance caused by fixed reward aggregation. MARBLE alleviates reward conflicts globally, and OmniNFT introduces modality-aware optimization, but neither adapts both reward coordination and update routing to the evolving model. Our method consistently improves modality quality, semantic consistency, cross-modal consistency, and synchronization, with the same trend across both backbones. \textbf{(iii) Training Dynamics.} Fig.~\ref{fig:reward}(a) shows that our method reaches the highest average reward while maintaining favorable trajectories across all five component rewards. In contrast, the baselines make less balanced progress across audio, video, and synchronization objectives. This indicates that the final gains arise from coordinated multi-reward optimization rather than improving one objective at the expense of others.

\begin{table*}[t]
  \centering
  \caption{\textbf{Results of the ablation studies on JavisBench using the LTX-2 backbone.} Gray-shaded rows isolate the reward-coordination components from the MARBLE baseline and do not inherit the routing stack. (Mean ± std over 3 seeds.)}
  \label{tab:ablation_ltx2}
  \resizebox{\textwidth}{!}{
  \begin{tabular}{ll cc cccc cc cc}
    \toprule
    \multirow{2}{*}{Study}
    & \multirow{2}{*}{Configuration}
    & \multicolumn{2}{c}{AV-Quality}
    & \multicolumn{4}{c}{Text-Consistency}
    & \multicolumn{2}{c}{AV-Consistency}
    & \multicolumn{2}{c}{AV-Synchrony} \\
    \cmidrule(lr){3-4}
    \cmidrule(lr){5-8}
    \cmidrule(lr){9-10}
    \cmidrule(lr){11-12}
    &
    & VQ$\uparrow$
    & AQ$\uparrow$
    & TV-IB$\uparrow$
    & TA-IB$\uparrow$
    & CLIP$\uparrow$
    & CLAP$\uparrow$
    & AV-IB$\uparrow$
    & AVHScore$\uparrow$
    & JavisScore$\uparrow$
    & DeSync$\downarrow$ \\
    \midrule

    \multirow{3}{*}{Routing}

    & + Token Weighting
    & 3.008\pstd{0.0127}
    & 5.663\pstd{0.0202}
    & 0.263\pstd{0.0018}
    & 0.156\pstd{0.0016}
    & 0.313\pstd{0.0013}
    & 0.388\pstd{0.0027}
    & 0.210\pstd{0.0022}
    & 0.201\pstd{0.0024}
    & 0.179\pstd{0.0019}
    & 0.482\pstd{0.0088} \\

    & + Layer Scale
    & 3.192\pstd{0.0111}
    & 5.784\pstd{0.0188}
    & 0.264\pstd{0.0017}
    & 0.161\pstd{0.0015}
    & 0.313\pstd{0.0012}
    & 0.409\pstd{0.0025}
    & 0.221\pstd{0.0020}
    & 0.227\pstd{0.0022}
    & 0.189\pstd{0.0018}
    & 0.366\pstd{0.0072} \\

    & + Token + Layer
    & 3.315\pstd{0.0103}
    & 5.839\pstd{0.0171}
    & 0.266\pstd{0.0015}
    & 0.162\pstd{0.0014}
    & 0.313\pstd{0.0011}
    & 0.411\pstd{0.0023}
    & 0.226\pstd{0.0018}
    & 0.229\pstd{0.0020}
    & 0.190\pstd{0.0016}
    & 0.343\pstd{0.0065} \\

    \rowcolor{gray!10}
    & + Branch-Aware
    & 2.612\pstd{0.0131}
    & 5.290\pstd{0.0211}
    & 0.264\pstd{0.0019}
    & 0.149\pstd{0.0017}
    & 0.311\pstd{0.0014}
    & 0.384\pstd{0.0030}
    & 0.198\pstd{0.0024}
    & 0.188\pstd{0.0027}
    & 0.177\pstd{0.0021}
    & 0.492\pstd{0.0096} \\

    \rowcolor{gray!10}
    & + Residual
    & 2.891\pstd{0.0118}
    & 5.526\pstd{0.0196}
    & 0.264\pstd{0.0016}
    & 0.158\pstd{0.0018}
    & 0.312\pstd{0.0013}
    & 0.393\pstd{0.0026}
    & 0.199\pstd{0.0021}
    & 0.199\pstd{0.0023}
    & 0.180\pstd{0.0019}
    & 0.455\pstd{0.0084} \\

    \rowcolor{gray!10}
    \multirow{-3}{*}{Weighting}
    & + Warm-Up
    & 2.999\pstd{0.0106}
    & 5.791\pstd{0.0179}
    & 0.265\pstd{0.0017}
    & 0.157\pstd{0.0014}
    & 0.312\pstd{0.0012}
    & 0.404\pstd{0.0024}
    & 0.209\pstd{0.0019}
    & 0.201\pstd{0.0021}
    & 0.183\pstd{0.0017}
    & 0.369\pstd{0.0070} \\

    \multicolumn{2}{c}{Routing + Weighting (Ours)}
    & \textbf{3.336}\pstd{0.0094}
    & \textbf{5.868}\pstd{0.0158}
    & \textbf{0.268}\pstd{0.0014}
    & \textbf{0.167}\pstd{0.0013}
    & \textbf{0.314}\pstd{0.0011}
    & \textbf{0.425}\pstd{0.0020}
    & \textbf{0.235}\pstd{0.0017}
    & \textbf{0.234}\pstd{0.0019}
    & \textbf{0.206}\pstd{0.0015}
    & \textbf{0.341}\pstd{0.0062} \\

    \bottomrule
  \end{tabular}
  }

\end{table*}

\subsection{Ablation Studies}
\label{sec:ablation}

\textbf{(i) Cross-Modal Influence-Guided Routing.} The routing ablation in Tab.~\ref{tab:ablation_ltx2} progressively introduces token weighting and layer scaling over GDPO. Token weighting improves local credit assignment by emphasizing tokens with stronger cross-modal responses, while layer scaling further improves consistency and synchronization by preserving gradients through influential interaction layers. Combining the two produces the strongest routing-only configuration, confirming that token- and layer-level adaptation are complementary. \textbf{(ii) Preference-Preserving Reward Coordination.} The gray rows isolate the weighting components from the MARBLE baseline without inheriting the routing stack. Branch-aware balancing assigns reward interactions to their responsible modality branches, while residual mixing preserves the predefined preference prior instead of replacing it with gradient-derived coefficients. Warm-up further stabilizes this adaptation by delaying dynamic reweighting until the estimated gradient relationships become reliable. \textbf{(iii) Complementarity and Reward Trade-Offs.} Individual components may favor different objectives, so intermediate configurations do not necessarily improve every metric monotonically. Nevertheless, progressively incorporating routing and weighting produces a stronger overall balance across quality, semantic consistency, and synchronization, as also reflected in Fig.~\ref{fig:reward}(b). Because the routing chain keeps the GDPO weighting fixed and the weighting chain inherits no routing, each chain isolates a single axis. The complete model performs best overall, demonstrating that adaptive update localization and preference-preserving reward coordination address distinct but complementary failure modes.

\subsection{Validating the Cross-Modal Influence Proxy}
\label{sec:proxy_validation}

\begin{figure}[t]
    \centering
    \includegraphics[width=\linewidth]{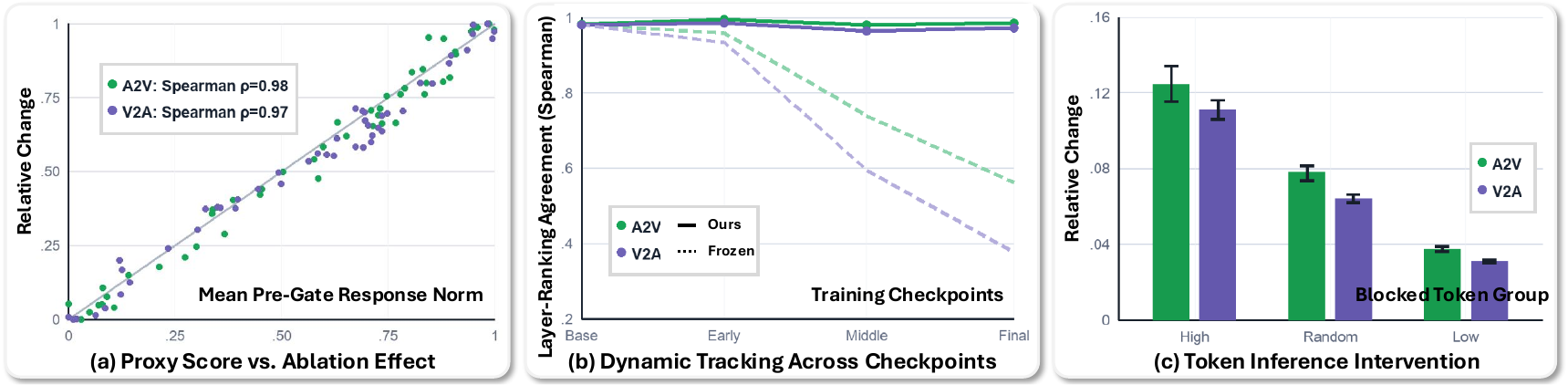}
    \caption{\textbf{The proxy correctly identifies influential cross-modal layers and tokens.}
    (a) Each point is one layer. The x-axis is the layer's mean pre-gate A2V/V2A response norm, while the y-axis is the relative change in the corresponding final video/audio velocity after disabling that layer. Both axes are min--max normalized across the 48 layers for visualization. Points near the diagonal indicate agreement between the proxy and direct intervention. (b) Higher correlation means better agreement with the current model. The current proxy stays accurate, whereas the proxy fixed at initialization becomes inaccurate as training changes the model. (c) We block the cross-modal information received by the top-scoring, random, or bottom-scoring 10\% of target tokens. A larger final prediction change means that the blocked tokens were more influential. The top-scoring tokens consistently cause the largest change, confirming that the proxy correctly identifies important tokens.}
    \label{fig:proxy_validation}
\end{figure}

We verify whether the response proxy identifies the cross-modal paths that actually affect the model output. At four training checkpoints, we disable each of the 48 A2V or V2A blocks separately while fixing the prompt, noisy latent, and timestep; a larger change in the final prediction indicates a more influential path. \textbf{(i) Layer-Level Fidelity:} The proxy closely recovers the intervention-based layer ranking, with Spearman correlations of $0.98$ for A2V and $0.97$ for V2A. Since a value close to $1$ means nearly identical rankings, the proxy reliably identifies which layers matter (Fig.~\ref{fig:proxy_validation}(a)). \textbf{(ii) Dynamic Tracking:} The proxy recomputed from the current model remains above $0.96$ throughout fine-tuning, whereas the proxy frozen at initialization falls to $0.56$ and $0.38$. Thus, influential layers shift during training, and a fixed routing map becomes stale (Fig.~\ref{fig:proxy_validation}(b)). \textbf{(iii) High-Score Tokens Matter More:} We block the cross-modal responses of the top-scoring, random, or bottom-scoring 10\% of target tokens. Blocking the top-scoring group changes the final prediction $1.62\times$ more for A2V and $1.74\times$ more for V2A than blocking an equally sized random group. Since the ablation size is identical, this result directly shows that higher proxy scores identify tokens with greater functional influence. 

\section{Conclusion}
\label{sec:conclusion}

This work reframes multi-reward post-training for joint audio-video generation as a dynamic credit-assignment problem. As optimization reshapes the model's cross-modal functions, both the token/layer localization of modality-conditioned updates and the relative strengths of reward signals should evolve accordingly. Adaptive Reward Routing embodies this principle by tracking changing cross-modal influence and reward conflicts throughout training, while preserving user-defined preferences. Its consistent gains across backbones, metrics, and ablations demonstrate the importance of adapting the optimization process in step with the model itself. More broadly, our findings point toward multimodal learning systems in which feedback is not routed by a fixed recipe, but continually reorganized as the model acquires new capabilities.

\paragraph{Limitations and Future Work.}
\textbf{(i) Reward Models:} No established unified reward jointly captures modality quality, semantic consistency, and temporal synchronization. Human-preference models~\citep{va-judger} provide a complementary overall signal but do not replace fine-grained, modality-specific supervision. Developing a comprehensive audio-video reward remains an important direction.
\textbf{(ii) RL Framework:} We adopt DiffusionNFT for stable, direct supervision of sampled flow-matching timesteps without reverse-process likelihood estimation. Our routing can extend to other diffusion objectives through their token losses and gradient paths, while reward coordination requires only reward-wise gradients.
\textbf{(iii) Architectural Scope:} We validate our method on both dual-stream LTX backbones and the unified single-stream JavisDiT++ backbone. Routing applies when modality tokens are identifiable and directional interaction responses can be isolated, whereas reward coordination is architecture-independent. Models with inseparable modality representations or inaccessible interaction responses remain future work.

\bibliography{iclr2027_conference}
\bibliographystyle{iclr2027_conference}

\end{document}